\documentclass[10pt]{article}

\usepackage{PRIMEarxiv}

\usepackage[utf8]{inputenc} 
\usepackage[T1]{fontenc}    
\usepackage{hyperref}       
\usepackage{url}            
\usepackage{booktabs}       
\usepackage{amsfonts}       
\usepackage{nicefrac}       
\usepackage{enumitem}
\usepackage{lipsum}
\usepackage{multirow}
\usepackage{lineno}
\usepackage{amsmath, amssymb, mathrsfs, bbm}
\usepackage{tabularx}
\usepackage{setspace}
\usepackage{subcaption}
\usepackage{xcolor}         
\usepackage{fancyhdr}       
\usepackage{graphicx}       
\usepackage[backend=biber,style=ieee,sorting=none,natbib=true]{biblatex}
\usepackage{authblk}
\usepackage{algorithm}
\usepackage{algpseudocode}
\usepackage{cleveref}
\usepackage{arydshln}

\usepackage{adjustbox}
\usepackage{changepage}
\usepackage{graphicx}
\usepackage{multirow}
\usepackage{booktabs}
\usepackage{lineno}
\usepackage{float}
\usepackage{soul}

\Crefname{figure}{Fig.}{Figs.} 
\Crefname{table}{Table}{Tables}
\Crefname{algorithm}{Alg.}{Algs.}
\Crefname{equation}{Eq.}{Eqs.}

\graphicspath{{media/}}     
\usepackage{bm}

\title{DINOv3 Visual Representations for Blueberry Perception Toward Robotic Harvesting}
\author[1]{Rui-Feng Wang}
\author[1]{Daniel Petti}
\author[2]{Yue Chen}
\author[1,*]{Changying Li}
\affil[1]{Bio-Sensing, Automation, and Intelligence Laboratory, Department of Agricultural and Biological Engineering, Institute of Food and Agricultural Sciences, University of Florida, Gainesville, FL 32611, USA}
\affil[2]{Department of Biomedical Engineering, Georgia Institute of Technology, Atlanta, GA 30332, USA}

\begin{document}
\maketitle

\let\thefootnote\relax
\noindent\footnotetext{$*$ Corresponding author: cli2@ufl.edu}

\begin{abstract}
Vision Foundation Models trained via large-scale self-supervised learning have demonstrated strong generalization in visual perception; however, their practical role and performance limits in agricultural settings remain insufficiently understood. This work evaluates DINOv3 as a frozen backbone for blueberry robotic harvesting-related visual tasks, including fruit and bruise segmentation, as well as fruit and cluster detection. Under a unified protocol with lightweight decoders, segmentation benefits consistently from stable patch-level representations and scales with backbone size. In contrast, detection is constrained by target scale variation, patch discretization, and localization compatibility. The failure of cluster detection highlights limitations in modeling relational targets defined by spatial aggregation. Overall, DINOv3 is best viewed not as an end-to-end task model, but as a semantic backbone whose effectiveness depends on downstream spatial modeling aligned with fruit-scale and aggregation structures, providing guidance for blueberry robotic harvesting. Code and dataset will be available upon acceptance.
\end{abstract}



\section{Introduction}
\label{sec:intro}
Reliable visual perception is a prerequisite for blueberry robotic harvesting under field conditions \citep{sun2025bmdnet}. In practice, harvesting strategies may target either individual fruits (“granular” harvesting) or spatially aggregated structures (“cluster” harvesting) \citep{gai2024hppem}, imposing distinct visual requirements. Region-level tasks such as fruit and bruise segmentation support quality assessment and post-harvest inspection, whereas instance-level fruit and cluster detection are critical for harvesting and yield estimation \citep{xu2021utilization, haydar2025integration}. Despite extensive model development, detection precision in field environments remains limited, particularly for dense clusters, and cross-condition robustness has yet to be achieved.

Recent Vision Foundation Models have demonstrated strong semantic generalization across visual domains \citep{lu2025vision,yan2025multimodal}. However, their practical role in harvesting-related perception tasks remains unclear. In frozen-backbone settings, patch discretization, spatial resolution, and task compatibility may impose structural constraints on downstream segmentation and detection performance. A systematic evaluation under harvesting-specific scenarios is therefore necessary. In this work, we evaluate DINOv3 (Self-Distillation with No Labels v3) \citep{simeoni2025dinov3} as a frozen visual representation backbone for blueberry perception tasks related to robotic harvesting. Under a unified experimental protocol, we analyze fruit and bruise segmentation as well as fruit and cluster detection across heterogeneous acquisition conditions. Our objectives are threefold: (1) to quantify task-specifc performance behavior under frozen representations; (2) to identify structural factors that constrain instance-level detection, particularly for small-scale and aggregation-defined targets; and (3) to clarify the operational role of foundation representations within harvesting perception systems.

\section{Related work}
\label{sec:related}
\subsection{Robotic harvesting-related visual perception task}
Visual perception for blueberry harvesting spans both region-level and instance-level formulations. For bruise segmentation, UNet3+ \citep{dai2025pebu} and YOLO-based approaches \citep{tan2025high} have been applied to extract damage regions. Fruit detection and segmentation have been addressed using supervised deep learning frameworks \citep{li2025field, singh2025development, zhao2025smart}, demonstrating feasibility under controlled conditions. 

However, performance often deteriorates under dense occlusion, scale variation, and cross-device acquisition shifts \citep{chen2024application}. In particular, cluster detection poses additional challenges because clusters are defined by spatial aggregation rather than single closed boundaries. Reported precision for cluster localization in field conditions remains limited to 40-50\% \citep{zhao2024object, li2025field}, reflecting the difficulty of modeling relational grouping structures. Existing studies primarily emphasize task-specific architectural refinement \citep{fu2025detection, huang2025identification, esaki2025maturity}, while systematic analysis of representational constraints and scale alignment issues remains scarce.

\subsection{Vision foundation models}
Vision Foundation Models trained via large-scale self-supervised learning provide transferable semantic representations with preserved spatial structure \citep{awais2025foundation}. DINOv3 produces dense patch-level embeddings that can serve as generic visual features \citep{simeoni2025dinov3}. Prior agricultural applications have explored foundation models through fine-tuning or adaptation strategies \citep{casas2025automated, wang2025optimized}. Nevertheless, limited work has examined their behavior as frozen backbones in harvesting-specific scenarios. In particular, the interaction between patch-level discretization and fruit-scale distributions, as well as the compatibility between generic representations and aggregation-defined targets, remains insufficiently understood. This gap motivates the present study.

\section{Methodology}
\subsection{Datasets}
To investigate the technical behavior and applicability of frozen DINOv3 representations for visual perception in robotic blueberry harvesting, we systematically evaluated blueberry datasets collected and curated by the Bio-Sensing, Automation \& Intelligence Lab (BSAIL, \url{https://uflbsail.net}) (Table \ref{tab:dinov3_datasets}). This study focuses on fruit- and cluster-level segmentation and detection tasks, aiming to characterize representation-level performance under field conditions and to provide technical insights for perception modules optimization in future harvesting systems.

\begin{table}[h]
\centering
\small
\caption{Datasets used for evaluating DINOv3.}
\label{tab:dinov3_datasets}
\begin{tabular}{l l r r r r r l}
\toprule
\textbf{Dataset} & \textbf{Total} & \textbf{Inst.} & \textbf{Train} & \textbf{Val.} & \textbf{Test} & \textbf{Ref.} \\
\midrule
Bruising Seg. & 1001 & 1642 & 799 & 101 & 101 & \citep{tan2025high} \\
Cluster Det. & 991 & 9361 & 855 & 60 & 76 & \citep{li2025field} \\
Fruit Det. & 1284 & 177602 & 898 & 256 & 130 & \citep{li2025field} \\
Fruit Seg. & 2772 & 129080 & 1940 & 554 & 278 & \citep{li2025field} \\
\midrule
\textbf{Total} & \textbf{6048} & \textbf{317685} & \textbf{4492} & \textbf{971} & \textbf{585} & -- \\
\bottomrule
\end{tabular}

\vspace{2mm}
\footnotesize
\textbf{Note:} “Total.”, “Train”, “Val,” and “Test” indicate number of images. “Inst.” is annotated objects.
\end{table}

\begin{figure}[htbp]
    \centering
    \includegraphics[width=0.7\linewidth]{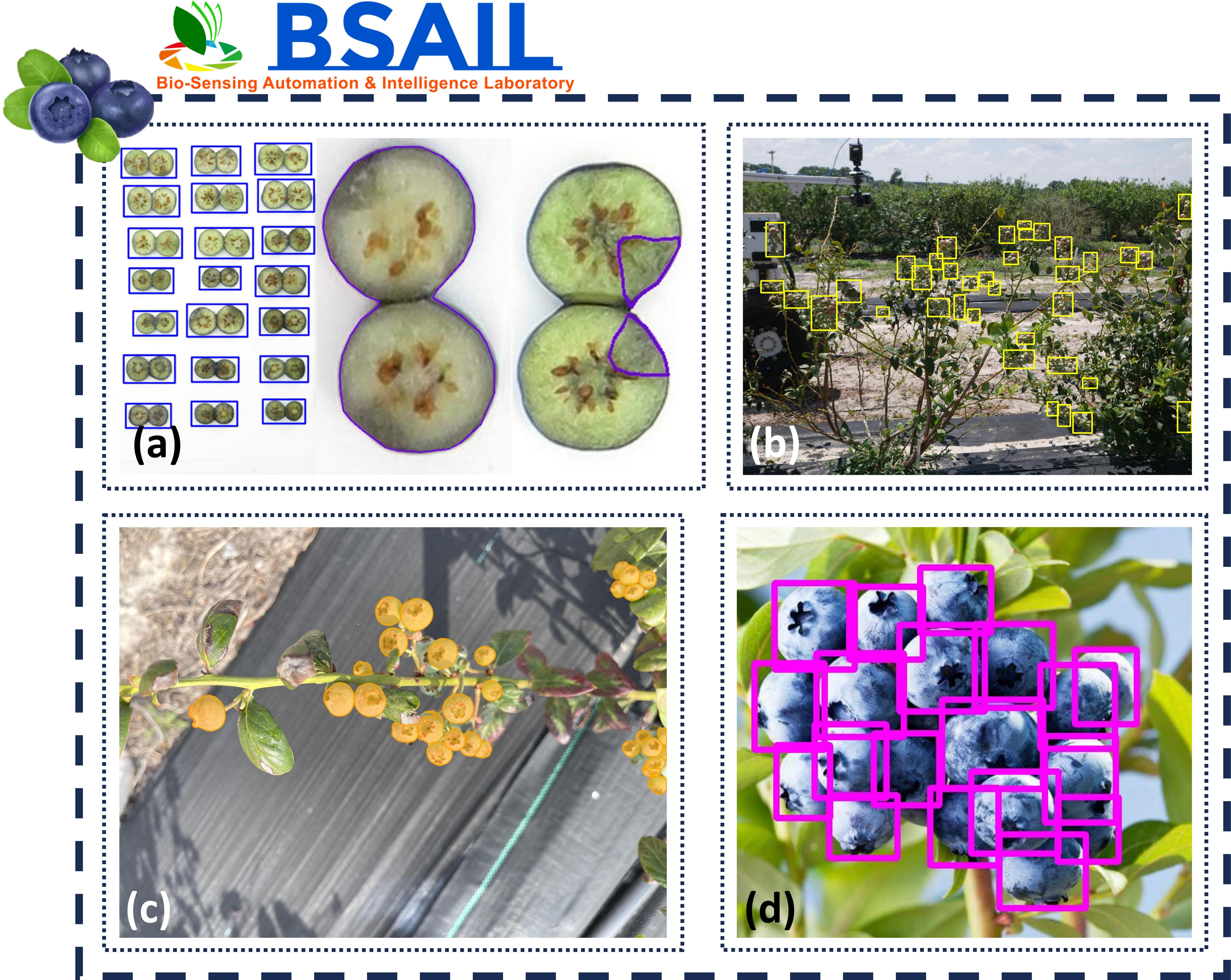}
    \caption{Samples of used datasets: \textbf{(a)} Bruising Segmentation; \textbf{(b)} Cluster Detection; \textbf{(c)} Fruit Segmentation; \textbf{(d)} Fruit Detection.}
    \label{fig:dataset_samples}
\end{figure}

The experimental analysis involves four blueberry-related datasets (Figure \ref{fig:dataset_samples}), covering fruit segmentation, fruit detection, cluster detection, and bruising segmentation. These datasets exhibit substantial variability in imaging conditions, object scale distributions, instance density, and occlusion patterns, reflecting the perceptual challenges encountered in practical harvesting scenarios. By evaluating frozen DINOv3 representations under a unified setting, we analyze their behavior across region-level semantic consistency and instance-level localization accuracy, with particular attention to task-dependent performance differences.


To ensure rigor and reproducibility, all datasets are standardized before training and evaluation. Datasets without predefined splits are randomly partitioned (7:2:1 for train/val/test), while original splits are retained when available. All annotations are converted to COCO-JSON format to maintain a consistent pipeline for feature extraction and downstream decoder training.

\subsection{Frozen DINOv3 evaluation framework}
To ensure a fair and reproducible evaluation of frozen DINOv3 representations, we construct a standardized feature extraction pipeline that unifies annotation normalization and backbone freezing within a single experimental framework. Given the heterogeneity of agricultural datasets \citep{cui2025efficient,wang2025sensors,zhang2026orchard,wang2024application}, all annotations are converted into a unified COCO-JSON format with consistent category definitions and geometric representations across detection and segmentation tasks. For datasets providing only segmentation masks \citep{zhang2025center,tan2025high}, bounding boxes (BBox) are automatically derived to maintain compatibility between region-level and instance-level supervision. This harmonized setup isolates representation quality as the primary source of performance variation, minimizing confounding factors from data formatting or training protocol inconsistencies while preserving the original semantic content of each dataset.

Following annotation normalization, DINOv3 models are employed as frozen visual encoders. We focus on four distilled variants [ViT-S/16 (21M parameters), ViT-S+/16 (29M), ViT-B/16 (86M), and ViT-L/16 (300M) \citep{simeoni2025dinov3}] selected to balance representational capacity and computational feasibility for agricultural deployment scenarios. All backbones remain fully frozen throughout training to decouple representation learning from task-specific optimization. This protocol allows us to isolate and analyze the intrinsic transferability and structural properties of DINOv3 features on unseen blueberry perception tasks, rather than assessing end-to-end task performance.

\begin{figure}[h]
    \centering
    \includegraphics[width=0.7\linewidth]{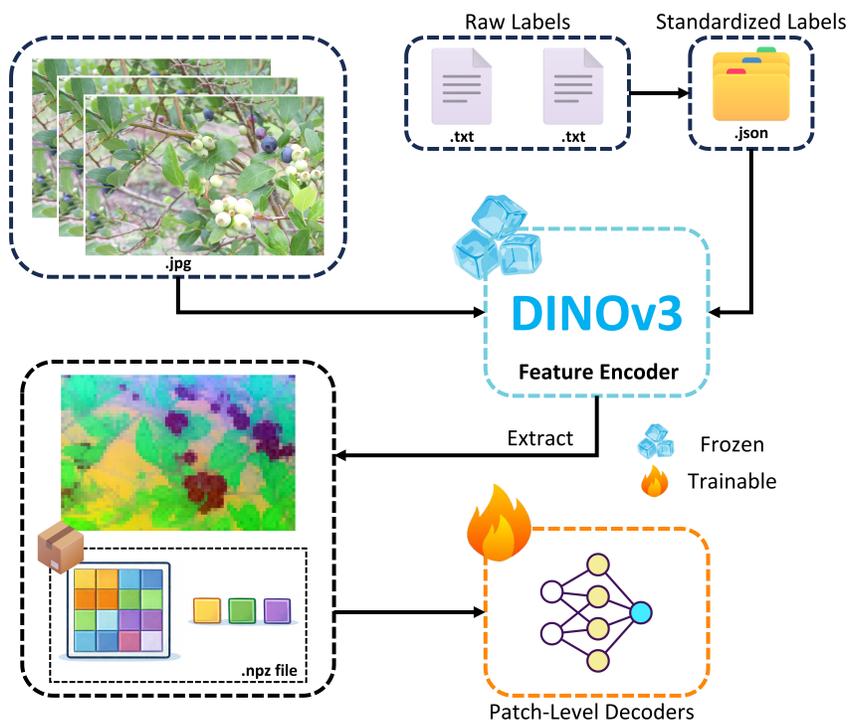}
    \caption{Workflow of DINOv3-based evaluation framework.}
    \label{fig:pipeline}
\end{figure}

For each input image, DINOv3 produces dense patch-level embeddings determined by a fixed $16 \times 16$ patch granularity, preserving spatial structure required for detection and segmentation. Features are precomputed and cached per subsets with the frozen DINOv3 encoder [compressed NumPy (.npz) format]. Using fixed representations for all head training and evaluation ensures fair comparisons across backbone variants and avoids redundant computation. These cached patch embeddings are the sole visual input to lightweight decoders for head-only training, quantitative evaluation, and qualitative visualization (Figure \ref{fig:pipeline}).

\subsection{Patch-level decoders}
\begin{figure}[h]
    \centering
    \includegraphics[width=0.7\linewidth]{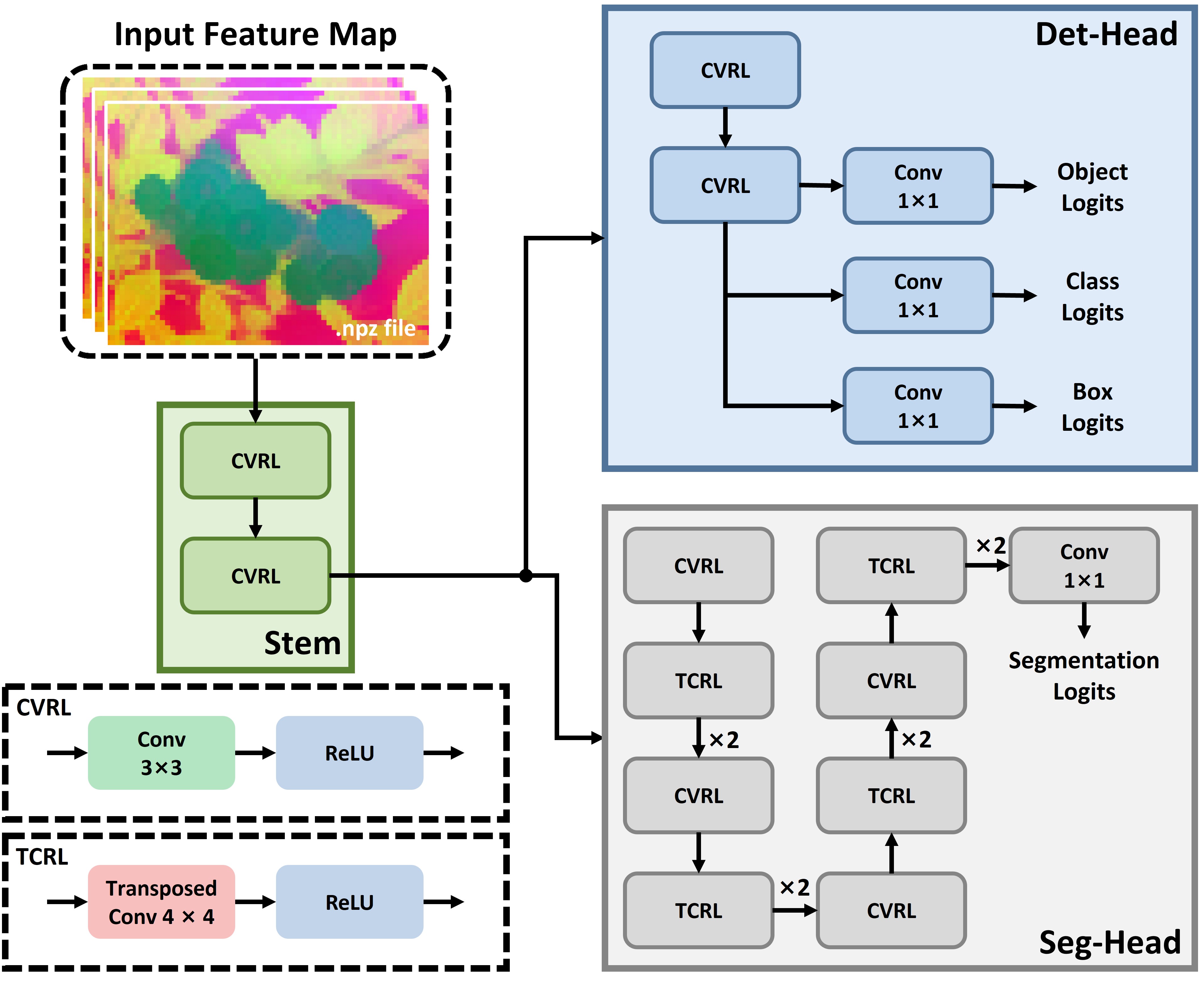}
    \caption{Structural diagram of Patch-Level Decoders.}
    \label{fig:decoder}
\end{figure}

Since DINOv3 operates as a frozen visual encoder and produces dense patch-level embeddings rather than task-specific predictions, downstream detection and segmentation must be formulated directly on the patch grid. The feature representations are defined as $\mathbf{F} \in \mathbb{R}^{B \times C \times H_p \times W_p}$, where $B$ denotes batch size, $C$ is channel dimension, and $(H_p, W_p)$ is spatial grid of patch tokens derived from non-overlapping $16 \times 16$ image regions. To adapt these representations for perception tasks, we employ lightweight task-specific decoders operating within the same patch coordinate system (Patch-Level Decoders, Figure \ref{fig:decoder}).

\textbf{Shared Adaptation Stem:}
To align frozen features with downstream objectives while preserving spatial structure, a lightweight adaptation module $\phi(\cdot)$ is first applied to yield a shared intermediate representation, $\mathbf{S} = \phi(\mathbf{F})$, where $\mathbf{S} \in \mathbb{R}^{B \times C_S \times H_p \times W_p}$ and $C_S$ denotes the adapted channel dimension. This shared representation provides a unified semantic space for both detection and segmentation branches, forming a typical hard parameter sharing framework \citep{caruana1993multitask}. By restricting learning to this lightweight module and subsequent heads, we isolate the contribution of DINOv3 representations from task-specific feature relearning, ensuring that performance differences primarily reflect representation quality rather than backbone optimization \citep{wang2023adaptive}.

\textbf{Patch-Level Detection Head (Det-Head):}
Object detection is formulated directly on the patch grid, where each patch serves as an implicit anchor aligned with the backbone’s token structure. Given the shared representation $\mathbf{S}$, the detection head predicts patch-level objectness ($\mathbf{O}\in\mathbb{R}^{B\times1\times H_p\times W_p}$), class scores ($\mathbf{C}\in\mathbb{R}^{B\times K\times H_p\times W_p}$) over $K$ categories, and bounding box offsets ($\mathbf{B}=(t_x,t_y,t_w,t_h)\in\mathbb{R}^{B\times4\times H_p\times W_p}$) defined in the local patch coordinate system. This anchor-free, patch-aligned design eliminates multi-scale heuristics and handcrafted anchor configurations, preserves geometric consistency with patch-level representations, and provides a structurally stable formulation for agricultural scenes characterized by sparse objects and local clustering.

\textbf{Patch-to-Pixel Segmentation Head (Seg-Head):}
Semantic segmentation requires dense pixel-level predictions. Since each patch corresponds to a fixed $P \times P$ image region ($P = 16$), the Seg-Head performs structured upsampling from the patch grid $(H_p, W_p)$ to the original image resolution $(H, W)$, satisfying $H = H_p \cdot P$ and $W = W_p \cdot P$. The decoder outputs a mask $\mathbf{M} = f_{\mathrm{seg}}(\mathbf{S})$, where $\mathbf{M} \in \mathbb{R}^{B \times 1 \times H \times W}$. Progressive upsampling stages are used to recover spatial detail while preserving alignment with the underlying patch structure. The additional layers in the segmentation decoder reflect resolution reconstruction requirements rather than increased semantic modeling capacity, ensuring fair comparison under the frozen-backbone setting.

\section{Experiments}
\subsection{Experimental setup}
All experiments are implemented in PyTorch and conducted on the HiperGator high-performance computing cluster. Each computing node was equipped with 8 AMD EPYC 9655P CPU cores, a single NVIDIA DGX B200 GPU with 180 GB of memory, and 32 GB of system RAM. DINOv3 backbones remain fully frozen, and only the adaptation layer and task-specific decoders are optimized. Models are trained for 50 epochs using the Adam optimizer with a learning rate of $1\times{10}^{-3}$ \citep{kingma2014adam}. Input images are resized while preserving aspect ratio, with dimensions adjusted to remain divisible by the patch size (16), and normalized using ImageNet statistics \citep{deng2009imagenet}. A lightweight data augmentation strategy consisting of horizontal flipping is applied during training. During inference, detection predictions are filtered using confidence thresholding and non-maximum suppression, and the maximum number of detections per image is capped to ensure stable and reproducible evaluation.

\subsection{Evaluation metrics}
Performance is evaluated using standard metrics for both semantic segmentation and object detection. While both tasks rely on true positives (TP), false positives (FP), and false negatives (FN), segmentation metrics are computed at the pixel level, whereas detection metrics are defined at the instance level under IoU-based matching.

For semantic segmentation, we report mean Intersection over Union (mIoU), Dice coefficient, Precision (P), and Recall (R) (Eq. \ref{eq:mIoU}-\ref{eq:P+R}). For object detection, P, R, and F1-score (F1) are computed at the instance level based on confidence thresholding and IoU matching, and further summarized performance using mean Average Precision (mAP) (Eq. \ref{eq:P+R}-\ref{eq:F1+map}). We report both mAP50 (IoU = 0.5) and mAP (IoU 0.50–0.95, step 0.05).

\begin{equation}\label{eq:mIoU}
    \mathrm{mIoU} = \frac{1}{K}\sum_{k=1}^{K} \frac{TP_{k}}{TP_{k} + FP_{k} + FN_{k}}
\end{equation}

\begin{equation}\label{eq:Dice}
    \mathrm{Dice} = \frac{2\text{TP}}{2\text{TP} + \text{FP} + \text{FN}}
\end{equation}

\begin{equation}\label{eq:P+R}
    \text{P}=\frac{\text{TP}}{\text{TP}+\text{FP}}, \quad
    \text{R}=\frac{\text{TP}}{\text{TP}+\text{FN}}
\end{equation}

\begin{equation}\label{eq:F1+map}
\text{F1}=\frac{2\text{P} \times \text{R}}{\text{P}+\text{R}},  \quad
    \text{mAP} = \frac{\sum_{n=1}^{N} \text{AP}(n)}{N}
\end{equation}

Here, $K$ is the number of semantic classes, AP($n$) is the Average Precision of class $n$, and $N$ is the number of object categories. All metrics are reported in percentage form.

\subsection{Quantitative results}\label{sec:quantitative}
Tables \ref{tab:seg_val} \& \ref{tab:det_val}, Tables \ref{tab:seg_test} \& \ref{tab:det_test} report the validation and test performance of different distilled DINOv3 variants on blueberry segmentation and detection tasks. Since all backbones remain frozen and only lightweight heads are trained, the results reflect the extent to which patch-level representations can be effectively read out under increasing model scale.

\textbf{Segmentation: Consistent Scaling Behavior.}
For semantic segmentation, performance improves consistently as the backbone scale increases from ViT-S to ViT-L. Both mIoU and Dice exhibit monotonic gains on validation and test sets (Tables \ref{tab:seg_val}-\ref{tab:seg_test}), indicating that stronger frozen representations translate into more stable region-level semantic coherence. Across fruit and bruising segmentation, larger backbones yield improved overlap quality and a more balanced P–R trade-off. These results suggest that when foreground regions are semantically coherent but exhibit local appearance variability, enhanced representational capacity can be effectively exploited by lightweight decoders to refine decision boundaries, even without backbone fine-tuning. Overall, segmentation shows stable and predictable scaling benefits under the frozen-representation paradigm.

\begin{table}[h]
\centering
\caption{Semantic segmentation results (\%) on the validation set.}
\label{tab:seg_val}
\small
\begin{tabular}{llccccc}
\toprule
Dataset & Model & mIoU & Dice & P & R \\
\midrule
\multirow{4}{*}{Bruising}
& ViT-S  & 67.470 & 79.330 & 83.189 & 80.122 \\
& ViT-S+ & 67.200 & 79.400 & 77.429 & \textbf{85.492} \\
& ViT-B  & 65.171 & 76.810 & \textbf{89.103} & 71.272 \\
& ViT-L  & \textbf{68.416} & \textbf{79.789} & 86.739 & 77.675 \\
\cmidrule(lr){1-6}
\multirow{4}{*}{Fruit Det.}
& ViT-S  & 66.564 & 78.438 & 79.944 & 78.169 \\
& ViT-S+ & 68.737 & 80.119 & 79.893 & 81.536 \\
& ViT-B  & 69.610 & 80.872 & 79.601 & 83.258 \\
& ViT-L  & \textbf{71.021} & \textbf{82.054} & \textbf{80.515} & \textbf{84.308} \\
\bottomrule
\end{tabular}
\end{table}

\begin{table}[h]
\centering
\caption{Semantic segmentation results (\%) on the test set.}
\label{tab:seg_test}
\small
\begin{tabular}{llccccc}
\toprule
Dataset & Model & mIoU & Dice & P & R \\
\midrule
\multirow{4}{*}{Bruising}
& ViT-S  & 66.487 & 78.861 & 77.255 & 83.860 \\
& ViT-S+ & 67.035 & 79.170 & 74.956 & \textbf{87.109} \\
& ViT-B  & 68.022 & 79.300 & \textbf{85.305} & 76.885 \\
& ViT-L  & \textbf{69.401} & \textbf{80.765} & 83.384 & 81.540 \\
\cmidrule(lr){1-6}
\multirow{4}{*}{Fruit Det.}
& ViT-S  & 64.847 & 77.282 & 78.249 & 77.845 \\
& ViT-S+ & 66.334 & 78.317 & 77.652 & 80.547 \\
& ViT-B  & 67.969 & 79.686 & 78.065 & 82.446 \\
& ViT-L  & \textbf{69.253} & \textbf{80.713} & \textbf{78.466} & \textbf{83.837} \\
\bottomrule
\end{tabular}
\end{table}

\textbf{Detection: Dataset-Dependent and Structurally Constrained.}
In contrast, object detection exhibits substantially greater heterogeneity in scaling effects (Tabes \ref{tab:det_val}-\ref{tab:det_test}). While larger backbones generally improve detection metrics for blueberry fruit detection, the gains are less uniform and more sensitive to target scale and spatial distribution. Notably, detection performance on the Blueberry Cluster Detection dataset remains extremely low across all backbone variants, with minimal improvement under scaling. This indicates that performance in such regimes is constrained more by spatial resolution and patch-level assignment mechanisms than by semantic representation capacity alone. Dense clustering, small object size, severe foreground–background imbalance at the patch level, or annotation definitions that are poorly aligned with bounding-box-based detection assumptions limit the effectiveness of patch-discretized localization, revealing a structural bottleneck independent of backbone scale.

\begin{table}[h]
\centering
\caption{Object detection results (\%) on the validation set.}
\label{tab:det_val}
\small
\begin{tabular}{llccccc}
\toprule
Dataset & Model & mAP50 & mAP & P & R & F1 \\
\midrule
\multirow{4}{*}{Cluster Det.}
& ViT-S  & 1.166 & 0.322 & \textbf{30.732} & 3.179 & 5.761 \\
& ViT-S+ & 1.244 & 0.305 & 23.304 & 3.986 & 6.807 \\
& ViT-B  & \textbf{1.770} & \textbf{0.483} & 29.050 & \textbf{5.247} & \textbf{8.889} \\
& ViT-L  & 0.750 & 0.203 & 24.348 & 2.825 & 5.063 \\
\cmidrule(lr){1-7}
\multirow{4}{*}{Fruit Det.}
& ViT-S  & 11.368 & 2.814 & 25.480 & 35.697 & 29.735 \\
& ViT-S+ & 13.016 & 3.301 & 21.876 & 39.244 & 28.092 \\
& ViT-B  & 11.210 & 2.391 & 20.879 & 34.926 & 26.135 \\
& ViT-L  & \textbf{16.031} & \textbf{4.444} & \textbf{29.290} & \textbf{42.800} & \textbf{34.779} \\
\cmidrule(lr){1-7}
\multirow{4}{*}{Fruit Seg.}
& ViT-S  & 24.348 & 7.005 & 36.741 & 50.196 & 42.427 \\
& ViT-S+ & \textbf{29.156} & \textbf{9.363} & 34.687 & 49.076 & 40.645 \\
& ViT-B  & 25.841 & 7.859 & 36.321 & \textbf{51.275} & 42.521 \\
& ViT-L  & 24.342 & 7.478 & \textbf{37.854} & 49.991 & \textbf{43.084} \\
\bottomrule
\end{tabular}
\end{table}

\begin{table}[h]
\centering
\caption{Object detection results (\%) on the test set.}
\label{tab:det_test}
\small
\begin{tabular}{llccccc}
\toprule
Dataset & Model & mAP50 & mAP & P & R & F1 \\
\midrule
\multirow{4}{*}{Cluster Det.}
& ViT-S  & 0.000 & 0.000 & 0.000 & 0.000 & 0.000 \\
& ViT-S+ & 0.000 & 0.000 & 0.000 & 0.000 & 0.000 \\
& ViT-B  & 0.000 & 0.000 & 0.000 & 0.000 & 0.000 \\
& ViT-L  & 0.000 & 0.000 & 0.000 & 0.000 & 0.000 \\
\cmidrule(lr){1-7}
\multirow{4}{*}{Fruit Det.}
& ViT-S  & 0.898 & 0.256 & 29.230 & 2.246 & 4.171 \\
& ViT-S+ & 0.844 & 0.216 & 24.839 & 2.765 & 4.977 \\
& ViT-B  & 1.814 & 0.375 & 33.528 & 5.387 & 9.282 \\
& ViT-L  & \textbf{2.166} & \textbf{0.576} & \textbf{37.568} & \textbf{5.757} & \textbf{9.985} \\
\cmidrule(lr){1-7}
\multirow{4}{*}{Fruit Seg.}
& ViT-S  & 0.451 & 0.082 & 29.118 & 1.520 & 2.889 \\
& ViT-S+ & \textbf{7.622} & \textbf{2.692} & \textbf{64.293} & \textbf{10.977} & \textbf{18.752} \\
& ViT-B  & 3.224 & 0.882 & 50.139 & 6.283 & 11.166 \\
& ViT-L  & 1.591 & 0.318 & 35.234 & 4.507 & 7.992 \\
\bottomrule
\end{tabular}
\end{table}

These observations highlight a fundamental distinction between segmentation and detection under frozen representations: segmentation benefits directly from improved semantic abstraction, whereas detection performance is jointly determined by representation quality and spatial discretization constraints imposed by the patch grid.

\textbf{Task-Level Divergence Under Identical Representations.}
A direct comparison of segmentation and detection on the same blueberry datasets further reveals task-dependent response patterns. Improvements in pixel-level overlap metrics do not always translate into proportional gains in box-level localization accuracy. Larger representations tend to increase recall-oriented behavior, which may benefit segmentation by expanding foreground coverage, yet lead to reduced precision in detection due to stricter instance-level matching criteria.

Overall, the quantitative results demonstrate that frozen DINOv3 representations provide consistent scalability for blueberry semantic segmentation, while detection performance is more strongly constrained by target density, object scale, and the inductive biases of patch-level localization.

\subsection{Qualitative analysis}
\begin{figure}[h]
    \centering
    \includegraphics[width=0.95\linewidth]{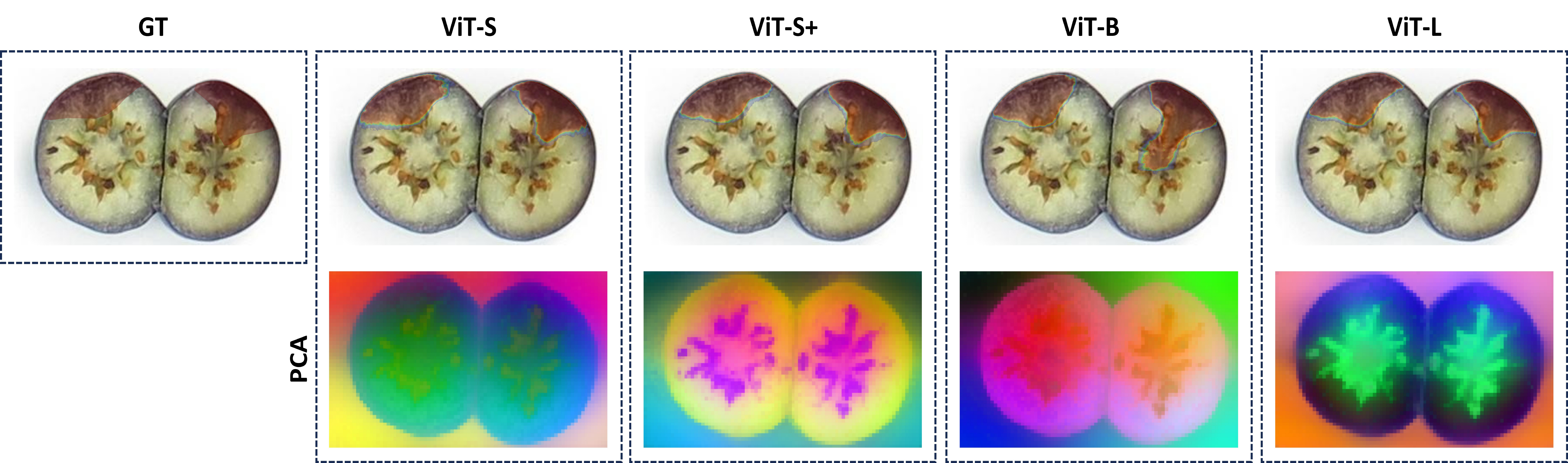}
    \caption{Visualization and PCA for bruising segmentation.}
    \label{fig:bruising}
\end{figure}

\begin{figure}[h]
    \centering
    \includegraphics[width=1.0\linewidth]{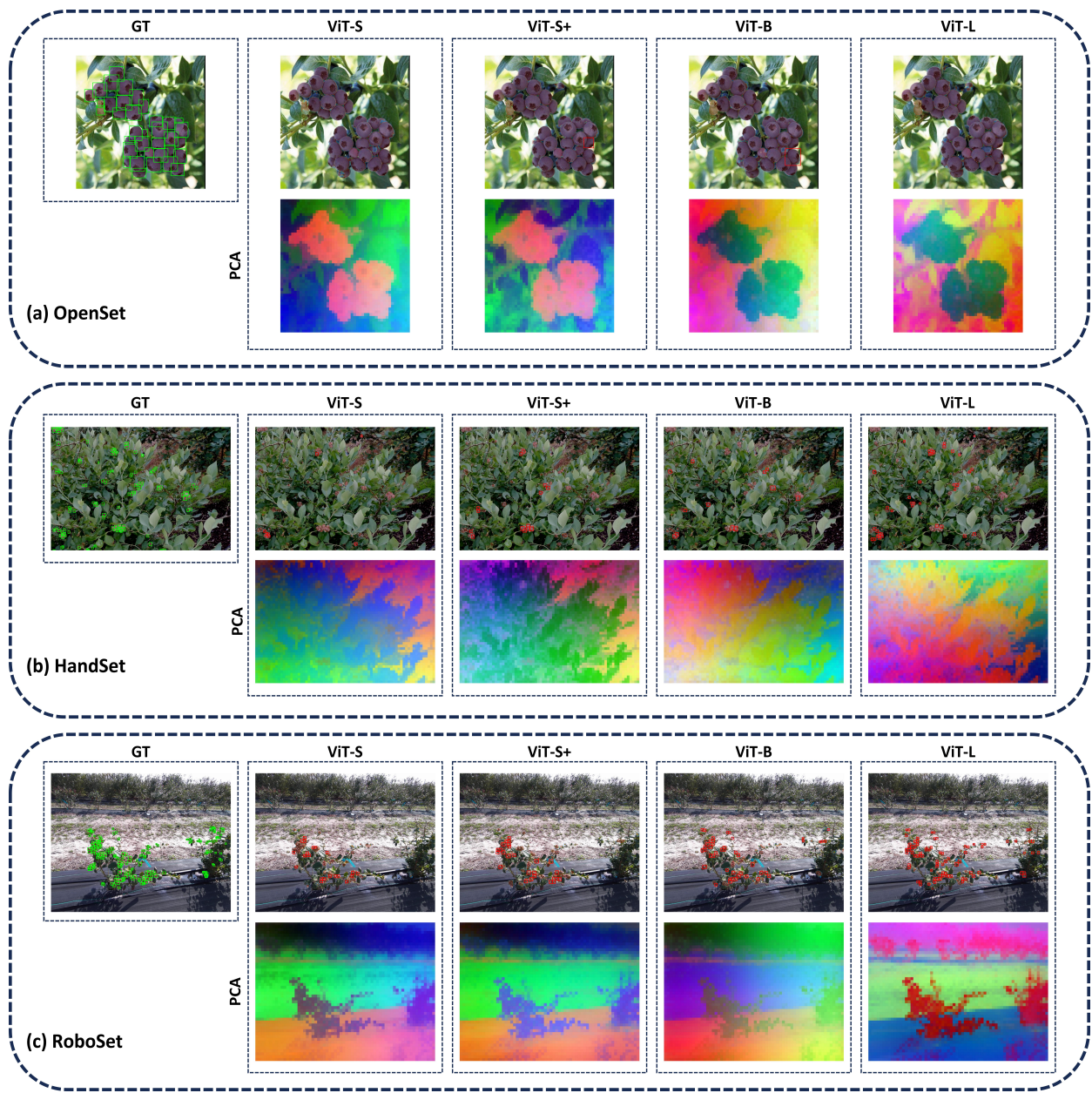}
    \caption{Segmentation and detection results with PCA analysis for the Blueberry Fruit Segmentation Dataset.}
    \label{fig:fruit_seg}
\end{figure}

\begin{figure}[h]
    \centering
    \includegraphics[width=1.0\linewidth]{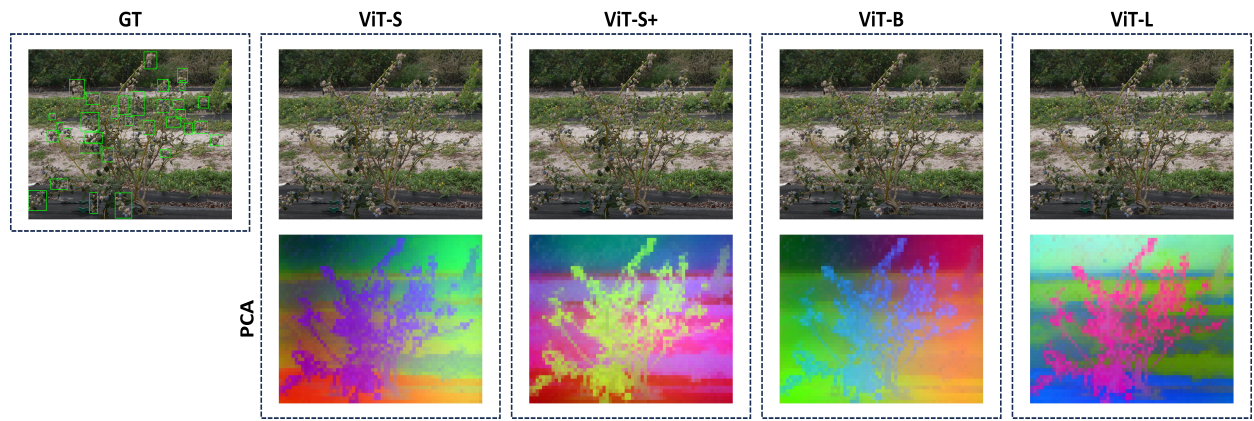}
    \caption{Visualization and PCA for cluster detection.}
    \label{fig:cluster}
\end{figure}

To complement the quantitative results and interpret model behavior from a representation perspective, we provide qualitative visualizations and PCA projections of patch-level features on the test datasets (Figures \ref{fig:bruising}–\ref{fig:cluster}). The qualitative evidence is consistent with the trends observed in Section \ref{sec:quantitative}: segmentation benefits steadily from backbone scaling, whereas detection remains more sensitive to spatial and structural constraints imposed by the patch grid.

\textbf{Segmentation: From Subtle Boundaries to Cross-Domain Stability.}
We first examine blueberry bruising segmentation (Figure \ref{fig:bruising}), a task characterized by subtle appearance variation and ambiguous boundaries. Although bruised regions are not clearly separable in low-dimensional PCA projections, larger DINOv3 variants still produce more coherent masks and improved overlap quality. This apparent discrepancy does not contradict the quantitative gains. PCA preserves only dominant variance directions, which typically capture global structural information rather than fine-grained texture variations. Prior studies have shown that perceptually subtle yet semantically meaningful differences may reside in phase- or frequency-related components that are not visually salient after dimensionality reduction \citep{oppenheim2005importance, geirhos2018imagenet}. Therefore, the absence of clear separation in PCA space does not imply weak discriminative capacity in the full high-dimensional representation.


We next consider fruit segmentation across three acquisition subsets [OpenSet (Figure \ref{fig:fruit_seg} (a)), HandSet (Figure \ref{fig:fruit_seg} (b), and RoboSet (Figure \ref{fig:fruit_seg} (c)] to assess cross-domain stability. In contrast to bruising, fruit-related patch embeddings form relatively compact clusters across subsets and remain separable from typical background components such as foliage and soil. These observations highlight a core property of DINOv3 as a frozen representation extractor: its dense patch-level features retain semantic consistency across acquisition domains, enabling a lightweight head to learn stable foreground decision boundaries from limited supervision. Importantly, the purpose of these visualizations is not to suggest that the backbone performs segmentation directly, but to demonstrate that, within a unified patch coordinate system, semantically consistent regions are mapped to nearby representations. This structural alignment reduces the effective learning burden of downstream decoders.


\textbf{Detection: Relational Targets and Spatial Bottlenecks.}
We begin with cluster detection, which represents the most structurally constrained regime. Across backbone variants, cluster detection collapses despite meaningful semantic structure observed in PCA projections (Figure \ref{fig:cluster}). It indicates that the limitation does not arise purely from weak feature separability. Instead, the bottleneck lies in the object definition itself. Unlike individual fruits, a “cluster” is not a single, closed-boundary object with stable geometry. Its definition depends on relational aggregation among multiple fruits under varying spatial configurations, occlusions, and discontinuities. Patch-level representations are naturally suited to capturing local semantic similarity but do not explicitly encode higher-order assembly relationships. Under a standard bounding-box detection formulation on a fixed patch grid, such relational targets become difficult to localize, as spatial grouping must be inferred implicitly from box regression rather than modeled directly. Consequently, performance is dominated by structural mismatch rather than representation capacity.


\begin{figure}[h]
    \centering
    \includegraphics[width=1.0\linewidth]{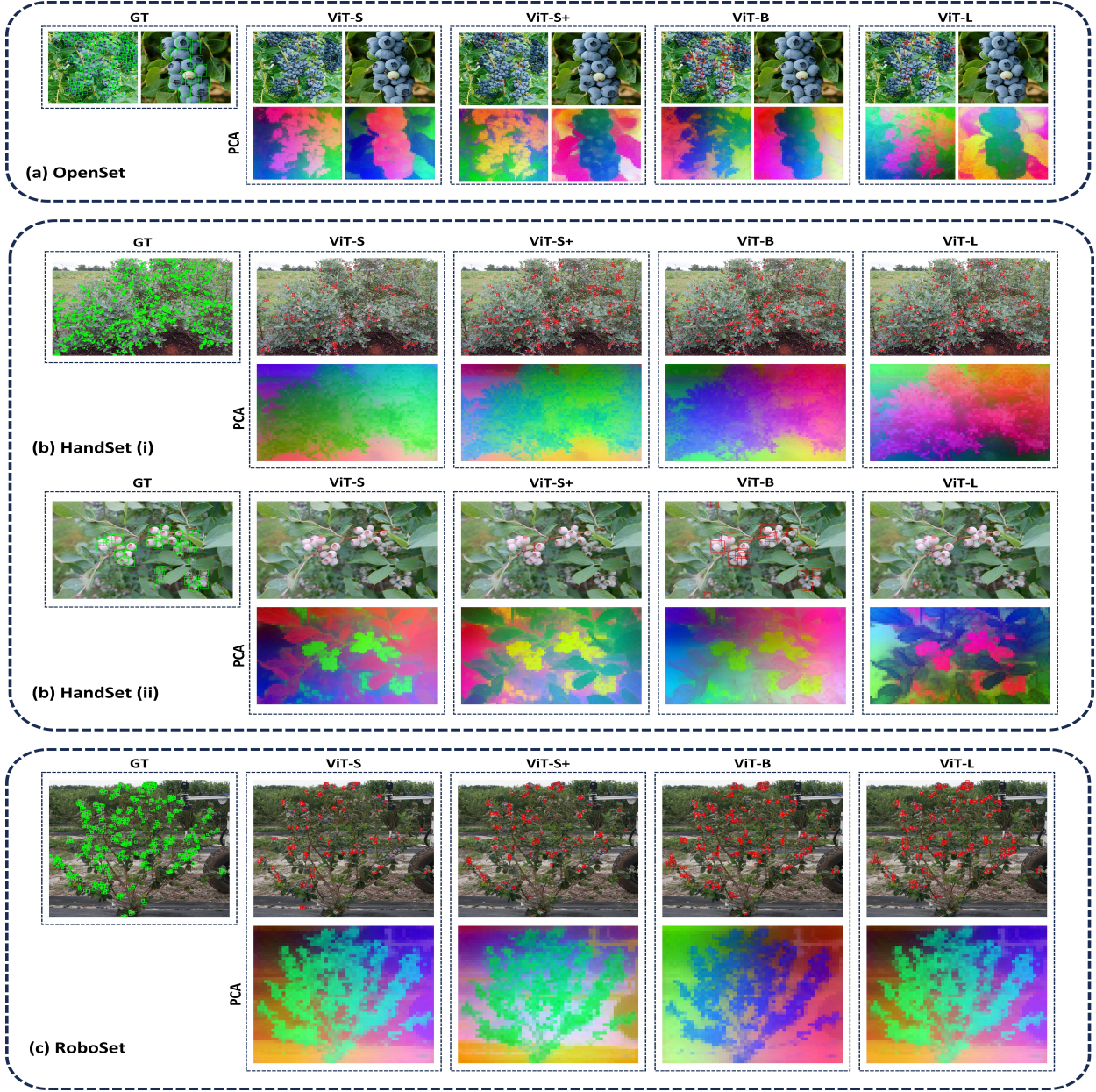}
    \caption{Visualization and PCA for the fruit detection.}
    \label{fig:fruit_det}
\end{figure}

In contrast, blueberry fruit detection exhibits partial scalability with backbone size, yet remains sensitive to spatial discretization. When fruit scale aligns more closely with patch granularity (OpenSet (Figure \ref{fig:fruit_det}(a) left), HandSet (Figure \ref{fig:fruit_det}(b)(i)), and RoboSet (Figure \ref{fig:fruit_det}(c))), semantic clusters in feature space translate into more stable proposals. However, when fruit extent spans multiple patches or is poorly aligned with the grid (OpenSet (Figure \ref{fig:fruit_det}(a) right) and HandSet (Figure \ref{fig:fruit_det}(b)(ii))), box localization remains unstable even if semantic separability is visually plausible. This mismatch reveals that under frozen representations, detection performance ceilings are jointly governed by semantic abstraction and the inductive biases of patch-level localization, including grid resolution, assignment strategy, and non-maximum suppression dynamics.


Therefore, detection results illustrate that while frozen DINOv3 representations provide strong semantic grouping at the patch level, successful instance-level localization requires compatibility between object definition, spatial granularity, and decoder formulation. Representation strength alone is insufficient when structural alignment is lacking.

Overall, the qualitative analyses reinforce the quantitative findings. Frozen DINOv3 representations maintain stable semantic structure across blueberry perception tasks, enabling reliable segmentation through lightweight readout. Detection performance, however, is governed not only by representation quality but also by structural alignment between patch granularity, target scale, and object definition. These observations clarify when frozen patch-level representations can be effectively leveraged and when performance bottlenecks arise from spatial formulation rather than semantic abstraction.

\section{Discussion}
\subsection{Technical insights}
This study clarifies the role of DINOv3 in blueberry visual perception toward robotic harvesting. Rather than functioning as an end-to-end segmentation or detection system, DINOv3 serves as a frozen general-purpose visual encoder that provides dense patch-level representations. The objective of this work is therefore not to compare DINOv3 against fully fine-tuned task models (e.g. SAM3 \citep{carion2025sam}, YOLO \citep{sapkota2025yolo}, UNet \citep{ronneberger2015u}), but to examine how effectively its frozen representations can be read out by lightweight decoders under harvesting-relevant perception tasks.

For bruising segmentation, scaling the backbone yields consistent improvements in region-level metrics despite the subtle appearance differences and ambiguous boundaries characteristic of bruise regions. Although low-dimensional PCA projections do not always visually isolate bruise patches, this does not contradict segmentation performance, as fine-grained texture variations may reside in lower-variance components not preserved in dominant projection axes. The results indicate that frozen DINOv3 features retain sufficient high-dimensional semantic structure to support stable foreground separation even when differences are not perceptually salient, suggesting that bruise quantitation in harvesting pipelines can benefit directly from stronger frozen representations with lightweight head-level adaption.

Fruit segmentation further demonstrates cross-domain stability across acquisition subsets, where patch embeddings corresponding to fruit regions remain semantically coherent despite variations in illumination, viewpoint, and sensor configuration. This consistency implies that region-level tasks relying on semantic compactness can be reliably supported under heterogeneous field conditions without full backbone fine-tuning, an important property for robotic harvesting systems in dynamic outdoor environments.

In contrast, detection exposes structural bottlenecks that are not resolved by increasing representation capacity alone. Instance-level localization is strongly affected by the spatial discretization imposed by the fixed patch grid, and a pronounced scale effect emerges when fruit size misaligns with patch granularity. When fruit extent spans multiple patches or varies substantially across acquisition regimes, bounding-box regression under a discretized grid cannot fully capture continuous spatial variation, leading to unstable localization despite preserved semantic separability. The failure of cluster detection reveals a deeper limitation: clusters are relational targets defined by spatial aggregation rather than single closed boundaries \citep{li2025field, gai2024hppem}, and fruit-level semantic separability does not automatically translate into cluster-level detectability without explicit grouping mechanisms. These observations indicate that detection performance ceilings in frozen-backbone settings might be aligned with spatial modeling strategies optimization (e.g. fruit-level aggregation priors \citep{gai2024hppem}, learnable grouping mechanisms within detection frameworks \citep{newell2017associative}, or graph-based reasoning \citep{hu2018relation}) rather than semantic abstraction strength.

From a harvesting-system perspective, improving blueberry detection reliability therefore requires aligning localization mechanisms with patch-level representations and fruit-scale distributions. Enhancements are more likely to arise from multi-scale feature fusion \citep{liu2025lightweight}, improved box regression under grid discretization \citep{zhang2024psrr}, adaptive assignment strategies \citep{dang2025adaptive}, or refined post-processing \citep{liu2026yolov5}, rather than from merely increasing backbone scale. Frozen DINOv3 representations provide a strong semantic foundation for blueberry perception, but effective deployment in robotic harvesting pipelines depends critically on explicitly engineered spatial reasoning that bridges semantic separability and instance-level localization.

\subsection{Future perspectives}
Building upon the empirical findings and technical insights presented in this study, future research on frozen DINOv3 representations for blueberry visual perception in robotic harvesting should focus on strengthening their integration into robotic harvesting pipelines rather than treating as standalone task models. The central implication of our results is that DINOv3 provides a stable semantic foundation, but effective harvesting-oriented perception requires architectural alignment between patch-level representations and task-specific spatial reasoning.

\begin{figure}[htbp]
    \centering
    \includegraphics[width=0.6\linewidth]{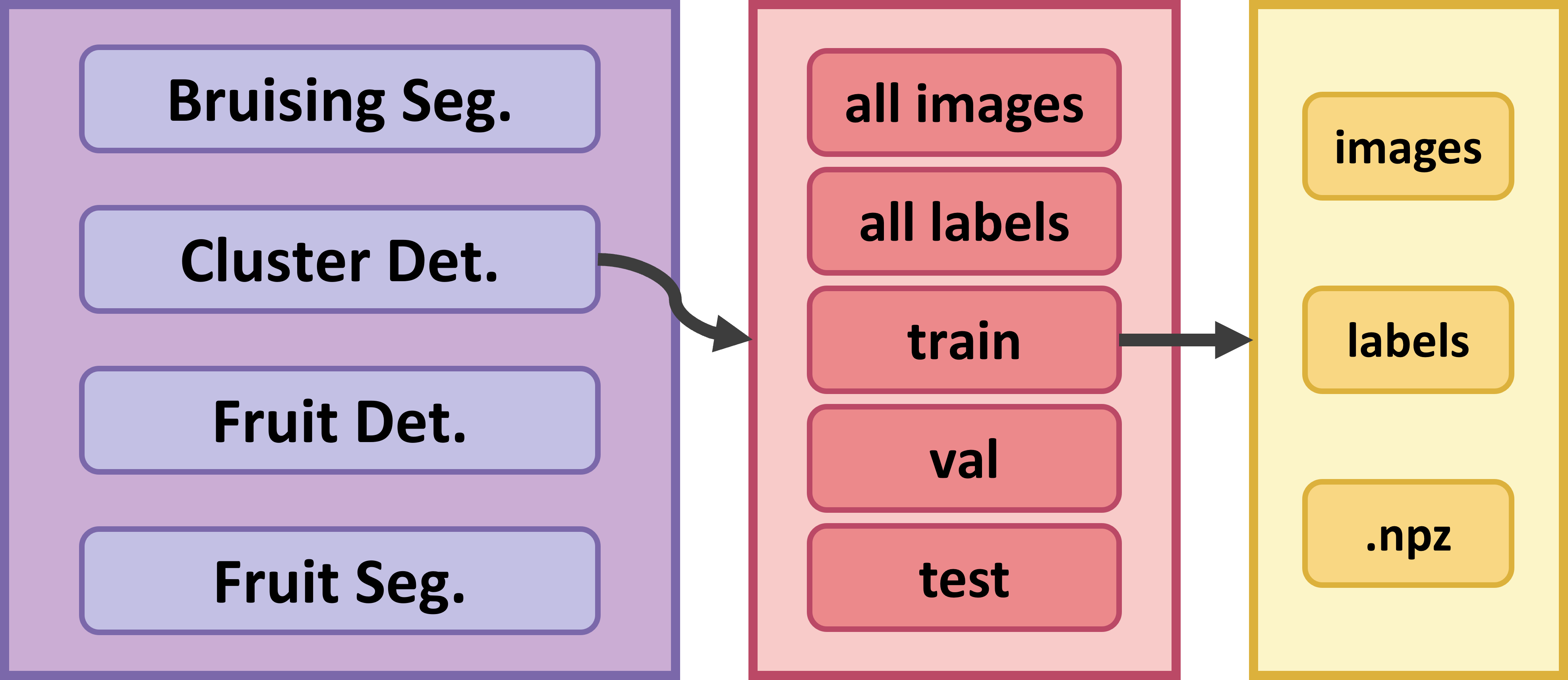}
    \caption{Structure of AgriSight-MT Dataset Blueberry section.}
    \label{fig:dataset_structure}
\end{figure}

A first practical direction lies in data infrastructure and reproducible evaluation. In this work, the datasets have been reorganized as a dedicated section within the AgriSight-MT dataset, accompanied by standardized splits and pre-extracted DINOv3 patch-level features (Figure \ref{fig:dataset_structure}). By publicly releasing this structured blueberry subset together with feature files and preprocessing utilities, we aim to enable controlled comparison of frozen-representation strategies under consistent protocols. For robotic harvesting research, such representation-centric infrastructure reduces overhead for experiments and reproducible evaluation, and facilitates systematic study of perception modules.

For region-level tasks, particularly fruit and bruise segmentation, future work may further explore lightweight readout mechanisms that fully exploit the semantic coherence preserved in frozen patch embeddings \citep{xue2025dinov3}. Instead of increasing network complexity, a promising paradigm for harvesting scenarios is to combine general-purpose representations with minimal, deployment-friendly decoders. This direction is especially relevant for field robots under limited computation budgets, where stable region extraction is often more critical than architectural complexity.

For instance-level detection, however, the primary bottleneck identified in this study is spatial rather than semantic. Future efforts should therefore emphasize better alignment between fruit-scale distributions and patch-grid discretization. Multi-scale feature aggregation, improved box regression across patch boundaries, and adaptive assignment mechanisms may help mitigate scale-induced localization instability. In harvesting contexts, where fruit size and density vary across viewpoints and growth stages, detection robustness will depend more on explicit spatial modeling than on further enlarging backbone capacity.

\begin{figure}[h]
    \centering
    \includegraphics[width=1.0\linewidth]{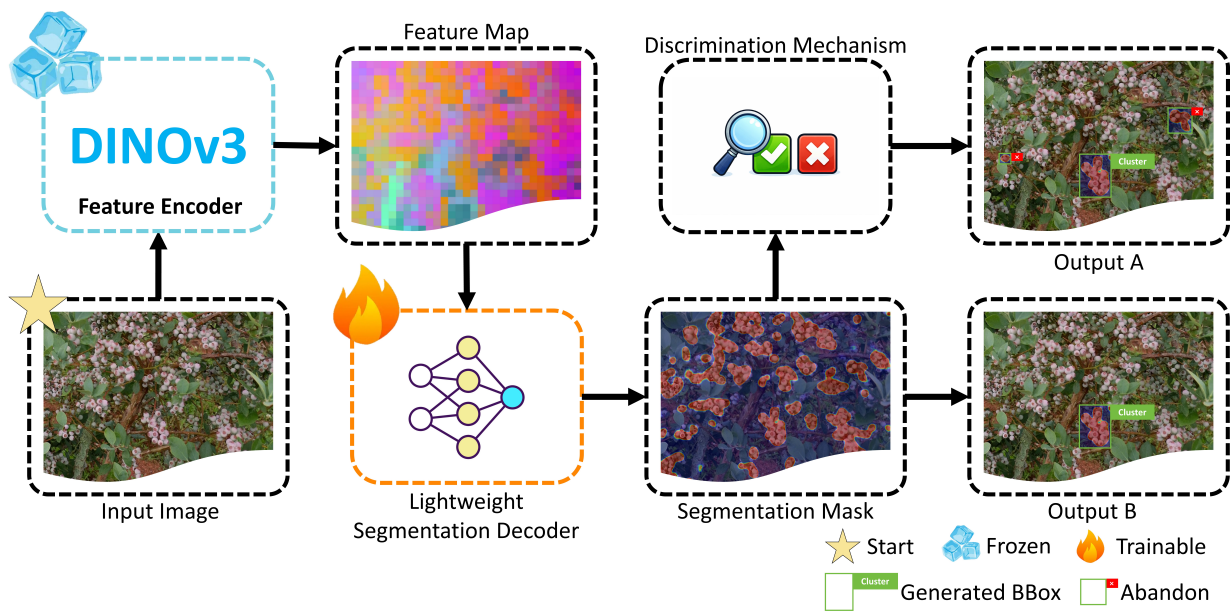}
    \caption{A conceptual pipeline for blueberry cluster detection based on fruit-level evidence aggregation. Output A generates cluster proposals by fitting BBoxes to predicted foreground regions and refines them using an aggregation verification module. The verification stage evaluates fruit-level cues within each ROI (e.g., fruit spatial compactness, connectivity) to filter proposals inconsistent with the cluster definition. Output B serves as a geometric baseline, where cluster boxes are obtained solely through direct region-to-box fitting without aggregation verification.}
    \label{fig:potential_cluster_det}
\end{figure}

The challenge becomes more pronounced for relational targets such as blueberry clusters. Because cluster identity is defined by aggregation structure rather than a single closed boundary, fruit-level separability in feature space does not automatically yield cluster-level detectability. The conceptual pipeline proposed in this study suggests elevating fruit-level evidence into group-level representations through lightweight aggregation verification (Figure \ref{fig:potential_cluster_det}). Future research may refine this paradigm by integrating learnable grouping mechanisms or structured reasoning modules that explicitly model spatial composition. Such strategies are likely essential for translating semantic cues into reliable cluster detection for harvesting and yield estimation.

Finally, although frozen DINOv3 features demonstrate semantic stability across acquisition subsets, detection remains sensitive to cross-domain scale shifts. Rather than relying solely on end-to-end adaptation, future work may explicitly incorporate scale-aware modeling strategies (such as multi-scale detection heads \citep{yang2026optimizing} or feature fusion \citep{huang2025real}) to handle viewpoint and resolution variability common in blueberry fields. Aligning semantic representation with scale-consistent spatial inference will be critical for deploying perception modules in real harvesting robots.

\section{Conclusions}
This study systematically evaluated DINOv3 as a frozen general-purpose visual encoder under a unified protocol for blueberry perception  tasks related to robotic harvesting. By designing task-aligned lightweight decoders and standardized training settings, we established a controlled framework to analyze its representation behavior across segmentation and detection. Results reveal clear task-dependent characteristics. Region-level segmentation benefits from stable and scalable dense patch representations, with consistent gains as backbone capacity increases. In contrast, instance-level detection is influenced not only by semantic separability but also by target scale distribution, patch discretization, and localization compatibility. Overall, these findings clarify the effective operating regime of frozen DINOv3 in harvesting perception. Rather than serving as an end-to-end model, it functions more appropriately as a semantic backbone, whose reliability depends on structural adaptation (multi-scale modeling \& relational reasoning) to translate strong semantic representations into robust instance-level localization and aggregation-aware detection.



\printbibliography 

@article{tan2025high,
  title={High throughput assessment of blueberry fruit internal bruising using deep learning models},
  author={Tan, Chenjiao and Li, Changying and Perkins-Veazie, Penelope and Oh, Heeduk and Xu, Rui and Iorizzo, Massimo},
  journal={Frontiers in Plant Science},
  volume={16},
  pages={1575038},
  year={2025},
  publisher={Frontiers Media SA}
}

@article{li2025field,
  title={In-field blueberry fruit phenotyping with a MARS-PhenoBot and customized BerryNet},
  author={Li, Zhengkun and Xu, Rui and Li, Changying and Munoz, Patricio and Takeda, Fumiomi and Leme, Bruno},
  journal={Computers and Electronics in Agriculture},
  volume={232},
  pages={110057},
  year={2025},
  publisher={Elsevier}
}

@article{zhang2026orchard,
  title={Orchard Chestnut Visual Harvest Maturity Detection and Segmentation Using an Improved YOLO-Based Method},
  author={Zhang, Yunhao and Zhang, Fan and Wang, Jiasheng and Yang, Hao and Zhang, Wenping and Li, Juan},
  journal={Agriculture},
  volume={16},
  number={4},
  pages={456},
  year={2026},
  publisher={MDPI}
}

@article{cui2025efficient,
  title={Efficient Localization and Spatial Distribution Modeling of Canopy Palms Using UAV Imagery},
  author={Cui, Kangning and Tang, Wei and Zhu, Rongkun and Wang, Manqi and Larsen, Gregory D and Pauca, Victor P and Alqahtani, Sarra and Yang, Fan and Segurado, David and Fine, Paul and others},
  journal={IEEE Transactions on Geoscience and Remote Sensing},
  year={2025},
  publisher={IEEE}
}

@article{wang2025sensors,
  title={From sensors to insights: Technological trends in image-based high-throughput plant phenotyping},
  author={Wang, Rui-Feng and Qu, Hao-Ran and Su, Wen-Hao},
  journal={Smart Agricultural Technology},
  pages={101257},
  year={2025},
  publisher={Elsevier}
}

@article{zhang2025center,
  title={Center-guided Classifier for Semantic Segmentation of Remote Sensing Images},
  author={Zhang, Wei and Ma, Mengting and Jiang, Yizhen and Lian, Rongrong and Wu, Zhenkai and Cui, Kangning and Ma, Xiaowen},
  journal={arXiv preprint arXiv:2503.16963},
  year={2025}
}

@article{simeoni2025dinov3,
  title={Dinov3},
  author={Sim{\'e}oni, Oriane and Vo, Huy V and Seitzer, Maximilian and Baldassarre, Federico and Oquab, Maxime and Jose, Cijo and Khalidov, Vasil and Szafraniec, Marc and Yi, Seungeun and Ramamonjisoa, Micha{\"e}l and others},
  journal={arXiv preprint arXiv:2508.10104},
  year={2025}
}

@inproceedings{caruana1993multitask,
  title={Multitask learning: A knowledge-based source of inductive bias1},
  author={Caruana, Rich},
  booktitle={Proceedings of the Tenth International Conference on Machine Learning},
  pages={41--48},
  year={1993}
}

@article{wang2023adaptive,
  title={Adaptive hard parameter sharing method based on multi-task deep learning},
  author={Wang, Hongxia and Jin, Xiao and Du, Yukun and Zhang, Nan and Hao, Hongxia},
  journal={Mathematics},
  volume={11},
  number={22},
  pages={4639},
  year={2023},
  publisher={MDPI}
}

@inproceedings{deng2009imagenet,
  title={Imagenet: A large-scale hierarchical image database},
  author={Deng, Jia and Dong, Wei and Socher, Richard and Li, Li-Jia and Li, Kai and Fei-Fei, Li},
  booktitle={2009 IEEE conference on computer vision and pattern recognition},
  pages={248--255},
  year={2009},
  organization={Ieee}
}

@article{kingma2014adam,
  title={Adam: A method for stochastic optimization},
  author={Kingma, Diederik P},
  journal={arXiv preprint arXiv:1412.6980},
  year={2014}
}

@article{oppenheim2005importance,
  title={The importance of phase in signals},
  author={Oppenheim, Alan V and Lim, Jae S},
  journal={Proceedings of the IEEE},
  volume={69},
  number={5},
  pages={529--541},
  year={2005},
  publisher={IEEE}
}

@inproceedings{geirhos2018imagenet,
  title={ImageNet-trained CNNs are biased towards texture; increasing shape bias improves accuracy and robustness},
  author={Geirhos, Robert and Rubisch, Patricia and Michaelis, Claudio and Bethge, Matthias and Wichmann, Felix A and Brendel, Wieland},
  booktitle={International conference on learning representations},
  year={2018}
}

@article{sapkota2025yolo,
  title={YOLO advances to its genesis: A decadal and comprehensive review of the You Only Look Once (YOLO) series},
  author={Sapkota, Ranjan and Flores-Calero, Marco and Qureshi, Rizwan and Badgujar, Chetan and Nepal, Upesh and Poulose, Alwin and Zeno, Peter and Vaddevolu, Uday Bhanu Prakash and Khan, Sheheryar and Shoman, Maged and others},
  journal={Artificial Intelligence Review},
  volume={58},
  number={9},
  pages={274},
  year={2025},
  publisher={Springer}
}

@inproceedings{ronneberger2015u,
  title={U-net: Convolutional networks for biomedical image segmentation},
  author={Ronneberger, Olaf and Fischer, Philipp and Brox, Thomas},
  booktitle={International Conference on Medical image computing and computer-assisted intervention},
  pages={234--241},
  year={2015},
  organization={Springer}
}

@article{carion2025sam,
  title={Sam 3: Segment anything with concepts},
  author={Carion, Nicolas and Gustafson, Laura and Hu, Yuan-Ting and Debnath, Shoubhik and Hu, Ronghang and Suris, Didac and Ryali, Chaitanya and Alwala, Kalyan Vasudev and Khedr, Haitham and Huang, Andrew and others},
  journal={arXiv preprint arXiv:2511.16719},
  year={2025}
}

@article{gai2024hppem,
  title={HPPEM: a high-precision blueberry cluster phenotype extraction model based on hybrid task cascade},
  author={Gai, Rongli and Gao, Jin and Xu, Guohui},
  journal={Agronomy},
  volume={14},
  number={6},
  pages={1178},
  year={2024},
  publisher={MDPI}
}

@inproceedings{hu2018relation,
  title={Relation networks for object detection},
  author={Hu, Han and Gu, Jiayuan and Zhang, Zheng and Dai, Jifeng and Wei, Yichen},
  booktitle={Proceedings of the IEEE conference on computer vision and pattern recognition},
  pages={3588--3597},
  year={2018}
}

@article{newell2017associative,
  title={Associative embedding: End-to-end learning for joint detection and grouping},
  author={Newell, Alejandro and Huang, Zhiao and Deng, Jia},
  journal={Advances in neural information processing systems},
  volume={30},
  year={2017}
}

@article{zhang2024psrr,
  title={PSRR-MaxpoolNMS++: Fast Non-Maximum Suppression With Discretization and Pooling},
  author={Zhang, Tianyi and Chen, Chunyun and Liu, Yun and Geng, Xue and Aly, Mohamed M Sabry and Lin, Jie},
  journal={IEEE Transactions on Pattern Analysis and Machine Intelligence},
  volume={47},
  number={2},
  pages={978--993},
  year={2024},
  publisher={IEEE}
}

@article{liu2025lightweight,
  title={A lightweight model based on multi-scale feature fusion for ultrasonic welding surface defect detection},
  author={Liu, Rui and Zhao, Lun and Ren, Yu and Shen, Zhonghua and Li, Liya and Luo, Jianfeng and Abbas, Zeshan},
  journal={Engineering Applications of Artificial Intelligence},
  volume={161},
  pages={112208},
  year={2025},
  publisher={Elsevier}
}

@article{dang2025adaptive,
  title={Adaptive spatial and scale label assignment for anchor-free object detection},
  author={Dang, Min and Liu, Gang and Chen, Chao and Wang, Di and Li, Xike and Wang, Quan},
  journal={Pattern Recognition},
  volume={164},
  pages={111549},
  year={2025},
  publisher={Elsevier}
}

@article{liu2026yolov5,
  title={YOLOv5-Based Dense Rice Seed Counting Method Integrating C3CBAM and Soft-NMS},
  author={Liu, Xiaoyang and Huang, Xupeng and Zhu, Rongjin and Hu, Chongyang and Wang, Cheng and Sun, Chenxin},
  journal={Smart Agricultural Technology},
  pages={101779},
  year={2026},
  publisher={Elsevier}
}

@inproceedings{xue2025dinov3,
  title={DINov3-UNet: A Pretrained Vision Transformer Enhanced U-Net for Surface Defect Segmentation},
  author={Xue, Tongjun and He, Zhenqian and Wei, Guanwei},
  booktitle={2025 5th International Symposium on Artificial Intelligence and Intelligent Manufacturing (AIIM)},
  pages={39--42},
  year={2025},
  organization={IEEE}
}

@article{huang2025real,
  title={Real-time object detection meets DINOv3},
  author={Huang, Shihua and Hou, Yongjie and Liu, Longfei and Yu, Xuanlong and Shen, Xi},
  journal={arXiv preprint arXiv:2509.20787},
  year={2025}
}

@article{yang2026optimizing,
  title={Optimizing seed anomaly detection in agricultural automation via lightweight ASD-YOLO and closed-loop control},
  author={Yang, Tianyu and Peng, Bo and Zhang, Jun and Zhang, Dongfang and Liang, Chenyang and Fan, XiaoFei},
  journal={Computers and Electronics in Agriculture},
  volume={243},
  pages={111342},
  year={2026},
  publisher={Elsevier}
}

@article{xu2021utilization,
  title={Utilization and development trend analysis of Vaccinium of America in blueberry breeding.},
  author={Xu, Guo-hui and Lei, Lei and An, Qi and Luo, Lin-qi and Wang, He-xin},
  journal={Journal of Fruit Science},
  volume={38},
  number={7},
  year={2021}
}

@article{haydar2025integration,
  title={Integration of machine learning models with real-time global positioning data to automate the wild blueberry harvester},
  author={Haydar, Zeeshan and Esau, Travis J and Farooque, Aitazaz A and Abbas, Farhat and Fraser, Andrew},
  journal={Precision Agriculture},
  volume={26},
  number={1},
  pages={7},
  year={2025},
  publisher={Springer}
}

@article{zhao2024object,
  title={Object detection in high-resolution UAV aerial remote sensing images of blueberry canopy fruits},
  author={Zhao, Yun and Li, Yang and Xu, Xing},
  journal={Agriculture},
  volume={14},
  number={10},
  pages={1842},
  year={2024},
  publisher={MDPI}
}

@article{dai2025pebu,
  title={PEBU-Net: A lightweight segmentation network for blueberry bruising based on Unet3+ using hyperspectral transmission imaging},
  author={Dai, Jingyuan and Wang, Guozheng and Yang, Mianqing and Liu, Dayang},
  journal={Measurement},
  volume={253},
  pages={117700},
  year={2025},
  publisher={Elsevier}
}

@article{singh2025development,
  title={Development of a smartphone application for rapid blueberry detection and yield estimation},
  author={Singh, Puranjit and Niknejad, Nariman and Spiers, James D and Bao, Yin and Ru, Sushan},
  journal={Smart Agricultural Technology},
  pages={101361},
  year={2025},
  publisher={Elsevier}
}

@article{zhao2025smart,
  title={Smart UAV-assisted blueberry maturity monitoring with Mamba-based computer vision},
  author={Zhao, Fan and He, Yinyin and Song, Jian and Wang, Jiaqi and Xi, Dianhan and Shao, Xinlei and Wu, Qingyang and Liu, Yongying and Chen, Yijia and Zhang, Guochen and others},
  journal={Precision Agriculture},
  volume={26},
  number={4},
  pages={56},
  year={2025},
  publisher={Springer}
}

@article{wang2025optimized,
  title={Optimized DINO model for accurate object detection of sesame seedlings and weeds},
  author={Wang, Yong and Xu, ShunFa and Ye, ZhenYuan and Cheng, KongHao},
  journal={Scientific Reports},
  volume={15},
  number={1},
  pages={11854},
  year={2025},
  publisher={Nature Publishing Group UK London}
}

@article{casas2025automated,
  title={Automated detection and segmentation of baby kale crowns using grounding DINO and SAM for data-scarce agricultural applications},
  author={Casas, Gianmarco Goycochea and Ismail, Zool Hilmi and Shapiai, Mohd Ibrahim and Karuppiah, Ettikan Kandasamy},
  journal={Smart Agricultural Technology},
  volume={11},
  pages={100903},
  year={2025},
  publisher={Elsevier}
}

@article{wang2024application,
  title={The application of deep learning in the whole potato production Chain: A Comprehensive review},
  author={Wang, Rui-Feng and Su, Wen-Hao},
  journal={Agriculture},
  volume={14},
  number={8},
  pages={1225},
  year={2024},
  publisher={MDPI}
}

@article{sun2025bmdnet,
  title={BMDNet-YOLO: A Lightweight and Robust Model for High-Precision Real-Time Recognition of Blueberry Maturity},
  author={Sun, Huihui and Wang, Rui-Feng},
  journal={Horticulturae},
  volume={11},
  number={10},
  pages={1202},
  year={2025},
  publisher={MDPI}
}

@article{lu2025vision,
  title={Vision foundation models in remote sensing: A survey},
  author={Lu, Siqi and Guo, Junlin and Zimmer-Dauphinee, James R and Nieusma, Jordan M and Wang, Xiao and Wernke, Steven A and Huo, Yuankai and others},
  journal={IEEE Geoscience and Remote Sensing Magazine},
  year={2025},
  publisher={IEEE}
}

@article{chen2024application,
  title={The application of optical nondestructive testing for fresh berry fruits},
  author={Chen, Zhujun and Wang, Juan and Liu, Xuan and Gu, Yuhong and Ren, Zhenhui},
  journal={Food Engineering Reviews},
  volume={16},
  number={1},
  pages={85--115},
  year={2024},
  publisher={Springer}
}

@article{fu2025detection,
  title={Detection of blueberry based on hyperspectral imaging and deep learning},
  author={Fu, Chengbiao and Liu, Siyi and Tian, Anhong},
  journal={Food Research International},
  pages={117141},
  year={2025},
  publisher={Elsevier}
}

@article{huang2025identification,
  title={Identification of early bruising degrees in blueberries using visible and near-infrared spectroscopy coupled with deep learning},
  author={Huang, Yuping and Bian, Zhouchen and Jin, Haojun and Zheng, Guoqing and Zhang, Qingyi and Hu, Dong and Xie, Weijun and Fan, Chenlong},
  journal={Spectrochimica Acta Part A: Molecular and Biomolecular Spectroscopy},
  pages={127200},
  year={2025},
  publisher={Elsevier}
}

@article{esaki2025maturity,
  title={Maturity Classification of Blueberry Fruit Using YOLO and Vision Transformer for Agricultural Assistance},
  author={Esaki, Ikuma and Noma, Satoshi and Ban, Takuya and Sultana, Rebeka and Shimizu, Ikuko},
  journal={Horticulturae},
  volume={11},
  number={10},
  pages={1272},
  year={2025},
  publisher={MDPI}
}

@article{yan2025multimodal,
  title={A multimodal vision foundation model for clinical dermatology},
  author={Yan, Siyuan and Yu, Zhen and Primiero, Clare and Vico-Alonso, Cristina and Wang, Zhonghua and Yang, Litao and Tschandl, Philipp and Hu, Ming and Ju, Lie and Tan, Gin and others},
  journal={Nature Medicine},
  volume={31},
  number={8},
  pages={2691--2702},
  year={2025},
  publisher={Nature Publishing Group US New York}
}

@article{awais2025foundation,
  title={Foundation models defining a new era in vision: a survey and outlook},
  author={Awais, Muhammad and Naseer, Muzammal and Khan, Salman and Anwer, Rao Muhammad and Cholakkal, Hisham and Shah, Mubarak and Yang, Ming-Hsuan and Khan, Fahad Shahbaz},
  journal={IEEE Transactions on Pattern Analysis and Machine Intelligence},
  volume={47},
  number={4},
  pages={2245--2264},
  year={2025},
  publisher={IEEE}
}

\end{document}